\documentclass[letterpaper, 10 pt, conference]{ieeeconf}
\IEEEoverridecommandlockouts
\usepackage{graphicx}
\usepackage[compress]{cite}
\usepackage{engord}
\usepackage{url}
\usepackage{multirow}
\usepackage{amsmath}
\usepackage{algorithm}
\usepackage{algorithmic}
\usepackage{float}
\usepackage{dsfont}
\usepackage[colorlinks=true, linkcolor=green, citecolor=green]{hyperref}
\usepackage{gensymb}
\usepackage{subfigure}
\usepackage{color,xcolor}
\definecolor{red}{HTML}{EA3323}
\definecolor{blue}{HTML}{00B0F0}
\definecolor{steer}{HTML}{000086}
\definecolor{acc}{HTML}{D2A741}

\title{\LARGE \bf
Multi-Task Conditional Imitation Learning for Autonomous Navigation at Crowded Intersections
}
\author{
	Zeyu~Zhu,
	Huijing~Zhao%
\thanks{Z.Zhu, H.Zhao are with the Key Lab of Machine Perception (MOE), Peking University, Beijing, China.}
\thanks{Contact: H.Zhao, zhaohj@pku.edu.cn.}
}

\begin{document}
\maketitle
\thispagestyle{empty} % no page number for the first page
\pagestyle{empty}  % no page number for the second and the later pages
\begin{abstract}
In recent years, great efforts have been devoted to deep imitation learning for autonomous driving control, where raw sensory inputs are directly mapped to control actions. However, navigating through densely populated intersections remains a challenging task due to uncertainty caused by uncertain traffic participants. We focus on autonomous navigation at crowded intersections that require interaction with pedestrians. A multi-task conditional imitation learning framework is proposed to adapt both lateral and longitudinal control tasks for safe and efficient interaction. A new benchmark called IntersectNav is developed and human demonstrations are provided. Empirical results show that the proposed method can achieve a success rate gain of up to 30\% compared to the state-of-the-art.
\end{abstract}
\section{Introduction}
Navigating through dense intersections is one of the most challenging tasks in autonomous driving due to the uncertainty created by pedestrians and other human-driven vehicles \cite{shu2020autonomous, shu2021driving,li2021planning,wei2021autonomous}.
 
 Nowadays autonomous agents generate driving policies at multiple levels of abstraction \cite{zhu2021survey}. Given a planned driving route on a map and mission points, during online navigation, rule- and modular-based methods are utilized to decide appropriate behavior and to plan trajectories taking into account the kinematic and dynamic constraints of the vehicle. MPC (model predictive control) \cite{camacho2013model,qian2015decentralized, schildbach2016collision} or PID (proportional-integral-derivative) \cite{misir1996design} finally realizes autonomous control. While these methods are easy to implement, they lack the ability to generalize in scenarios that cannot be accurately modeled beforehand. Therefore, in this case, the parameters are tuned to guarantee safety first. At dense intersections, today's autonomous driving systems are often complained of conservative behavior, inefficiency and inhuman driving.

Recently, great efforts in deep learning-based autonomous driving control have been witnessed \cite{kuutti2020survey, zhu2021survey}. The appeal of deep learning is that sensorimotor control actions can be implicitly learned from sensory input (e.g., front-view images) in an end-to-end fashion, where deep reinforcement learning (DRL) \cite{bouton2019safe, liang2018cirl, zhang2021end} and deep imitation learning (DIL) \cite{bojarski2016end, codevilla2018end,codevilla2019exploring, zhao2019sam, chen2020learning} are two representatives. On the one hand, DRL typically learns from online trial and error (i.e., interaction with the environment), which can be dangerous in real world. Therefore, most current DRL methods \cite{bouton2019safe, liang2018cirl, zhang2021end} rely heavily on simulators. On the other hand, DIL learns from expert demonstrations and can be executed offline, which is important for safety-critical applications such as autonomous driving \cite{codevilla2018end,codevilla2019exploring}. Furthermore, it has the potential to achieve human-like driving through human demonstrations, which can be easily collected using low-cost on-board sensors.
 
 Despite recent success, deep imitation learning still suffers from covariate shift \cite{ross2010efficient} and causal confusion \cite{de2019causal}.
  Generalizing to dense traffic scenarios (e.g., intersections with many pedestrians) remains an open problem \cite{codevilla2019exploring}, where autonomous agents need to perform both lateral and longitudinal controls simultaneously to interact with pedestrians on crosswalks, and navigate the intersection safely and efficiently. 
While some DIL studies have shown results for intersection navigation \cite{codevilla2018end, sauer2018conditional, codevilla2019exploring}, control strategies when interacting with pedestrians have not been rigorously studied, and the different nature of lateral and longitudinal control has been ignored. Several DIL benchmarks \cite{dosovitskiy2017carla,codevilla2019exploring,carlaleaderboard} were developed on a high-fidelity CARLA simulator \cite{dosovitskiy2017carla} in urban scenes. However, none of them focused on intersection navigation or interaction with pedestrians, which may be a reason for limiting research.

This study investigates DIL-based autonomous driving policy learning for intersection navigation with pedestrian interaction. To address the different uncertainty in lateral and longitudinal control, a multi-task setup based on a homoscedastic uncertainty is designed. By extending the popular Conditional Imitation Learning (CIL) \cite{codevilla2018end} framework, this work proposes Multi-Task Conditional Imitation Learning (MTCIL) to adapt lateral and longitudinal control simultaneously for safe and smooth interaction with pedestrians on crosswalks, meanwhile navigating through intersections efficiently. A new benchmark called IntersectNav is developed, in which about 800 human driving trajectories on 40 routes are collected at four intersections under different weather conditions for train and validation. The other two intersections are used for testing. In addition, new evaluation protocols and metrics are defined to enrich the criteria of traditional benchmarks. The performance of the proposed method is extensively studied, where experimental results show that our model achieves up to 30\% success rate gain compared to the state-of-the-art. The benchmark, collected dataset and video are available at https://github.com/zhackzey/IntersectNav.

Our paper is organized as follows. Section \ref{sec:related work} related work. Section \ref{sec:methodology} the proposed method. Section \ref{sec:benchmark} the proposed benchmark. Section \ref{sec:experiment} experimental results and Section \ref{sec:conclusion} our conclusion.

\section{Related Work}
\label{sec:related work}
\subsection{Visual-based Imitation Learning for Autonomous Driving}

\textit{Direct perception methods} \cite{chen2015deepdriving,sauer2018conditional} utilize neural networks to extract compact intermediate representations which are then passed to subsequent decision and control modules. CAL \cite{sauer2018conditional} learns to predict affordances, such as distance to the preceding vehicle. However, affordance design requires system expertise, which may not be optimal. 

\textit{End-to-end methods} \cite{pomerleau1988alvinn, bojarski2016end, codevilla2018end} learn to map raw sensor input (e.g., images) to control signals (e.g., acceleration, steering). Bojarski et.al \cite{bojarski2016end} successfully learned a steering policy. However, their model only adapts to lane keeping and has difficulty in addressing complex scenarios. Codevilla et.al proposed Conditional Imitation Learning (CIL) \cite{codevilla2018end}, where the output is conditioned on high-level commands. They also proposed CILRS \cite{codevilla2019exploring}, an improved version of CIL. However, these models have limitations in generalizing to dense traffic due to the instinctive covariate shift problem \cite{ross2010efficient} of imitation learning, . Furthermore, offline imitation learning suffers from causal confusion \cite{de2019causal}, where the model cannot distinguish spurious correlations from true causes in observed training demonstration patterns. A large body of CIL-based work has been proposed to address these issues. 

Privileged supervisions such as road maps (LBC \cite{chen2020learning}) or BEV representations (Roach \cite{zhang2021end}) are used as input. Object-level detections such as vehicles and pedestrians can be integrated into the input, reducing the perceptual burden on DNNs compared to front-view images. Although privileged information can be easily and efficiently accessed in the simulator, retrieving it from real-world observations is not trivial. To overcome the covariate shift problem, some works \cite{prakash2020exploring, chen2020learning} employ DAgger \cite{ross2010efficient} to transfer offline imitation learning to online refinement. Alternatively, online/on-policy reinforcement learning is utilized for more exploration, where an offline trained IL agent serves as the initialization of the RL agent (CIRL \cite{liang2018cirl}, LSD \cite{ohn2020learning}), or the IL agent imitates a well-trained RL agent (Roach \cite{zhang2021end}). However, both DAgger and online RL can only perform effectively in simulations because accessing online demonstrations in real-world is not trivial. They also suffer from expensive training costs. Besides, a well-designed reward function is crucial for the learned policy \cite{zhu2021survey}, which may not reflect realistic human driving behavior.
Our work differs from the above works in several ways. First, we leverage efficient offline imitation learning and introduce an additional longitudinal branch to overcome casual confusion, where a simulator is unnecessary. Therefore, our work is more scalable to the real world. Second, we focus on more complex interactive traffic scenarios. Third, we learn from human's rather than autopilot agents' demonstrations in previous work.

\subsection{Multi-task Learning in Computer Vision}
Multi-task learning \cite{ruder2017overview, zhang2021survey} aims to improve learning efficiency by learning multiple complimentary tasks from shared representations. Many multi-task methods have been proposed for computer vision. For semantic tasks, different combinations of tasks can be used (e.g., classification and semantic segmentation \cite{liao2016understand} or detection \cite{sermanet2013overfeat}). For geometry and regression tasks, depth, surface normals and semantic segmentation are learned in \cite{eigen2015predicting}.

Some works build on multi-task learning and learn policies for robotics \cite{xu2018shared} or self-driving \cite{xu2017end, kim2020multi, ishihara2021multi}. Kim et.al \cite{kim2020multi} used prediction of future actions and states as side tasks and learned together with primary control task in multi-task learning fashion. \cite{xu2017end, ishihara2021multi} trained the policy together with semantic segmentation side task to obtain a meaningful and generic feature space. Our method differs from these methods in several ways. Instead of introducing side tasks that increases training cost, we split the primary task into lateral and longitudinal tasks, which are learned together in multi-task setting. Since the units and scales of two tasks are different, we build upon homoscedastic uncertainty and learn to adjust their weights adaptively. Empirical results demonstrate our effectiveness.

\begin{figure}[h]
	\centering
	\includegraphics[width=0.95\linewidth]{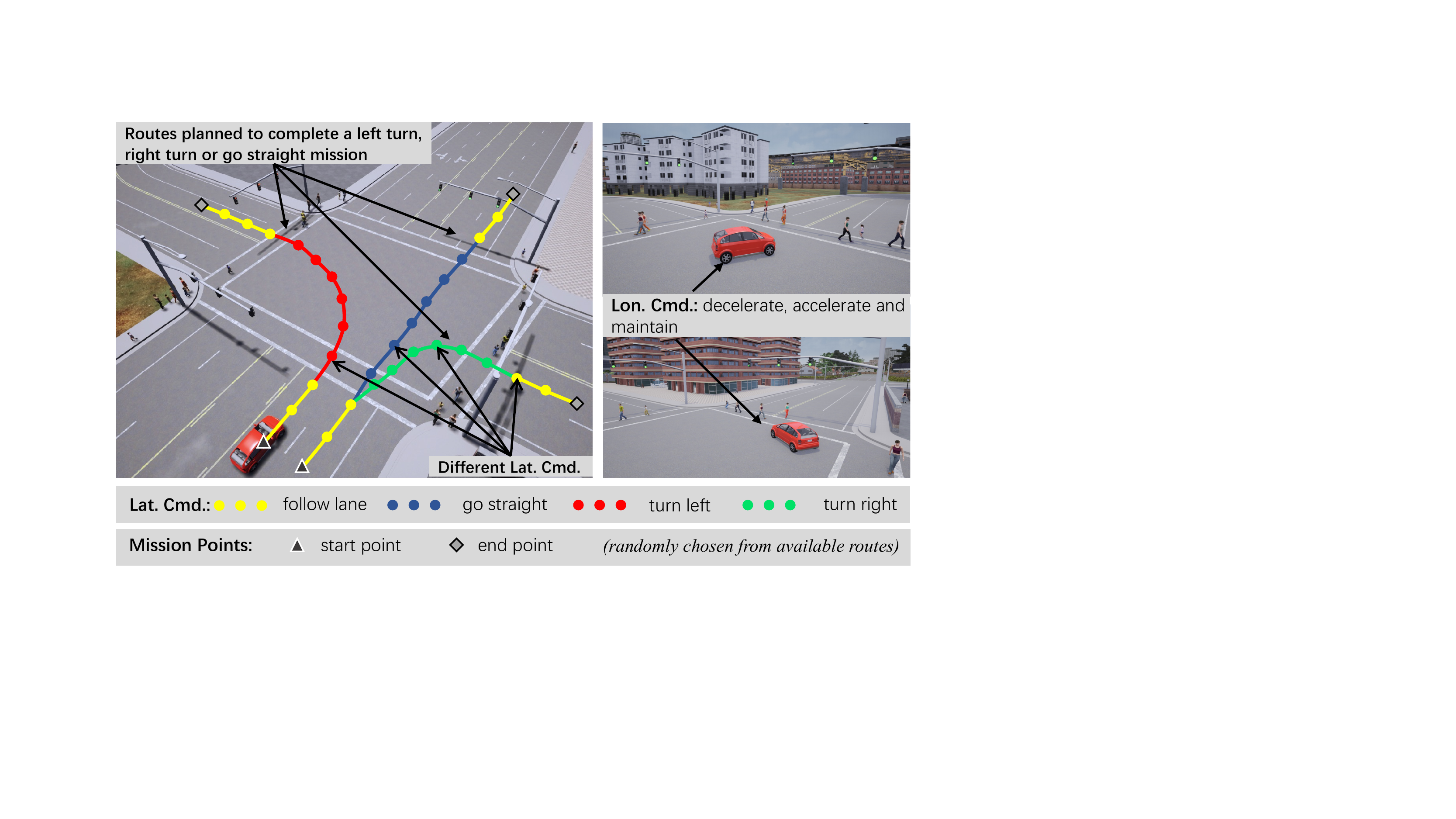}
	\caption{Illustration of intersection scenarios. Given a planned route and high-level commands, the agent needs to complete three kinds of missions.}
	\label{fig:scene_illustration}
%	\vspace{-6mm}
\end{figure}
\begin{figure*}[ht]
\vspace{3mm}
\centering
\includegraphics[width=0.85\textwidth]{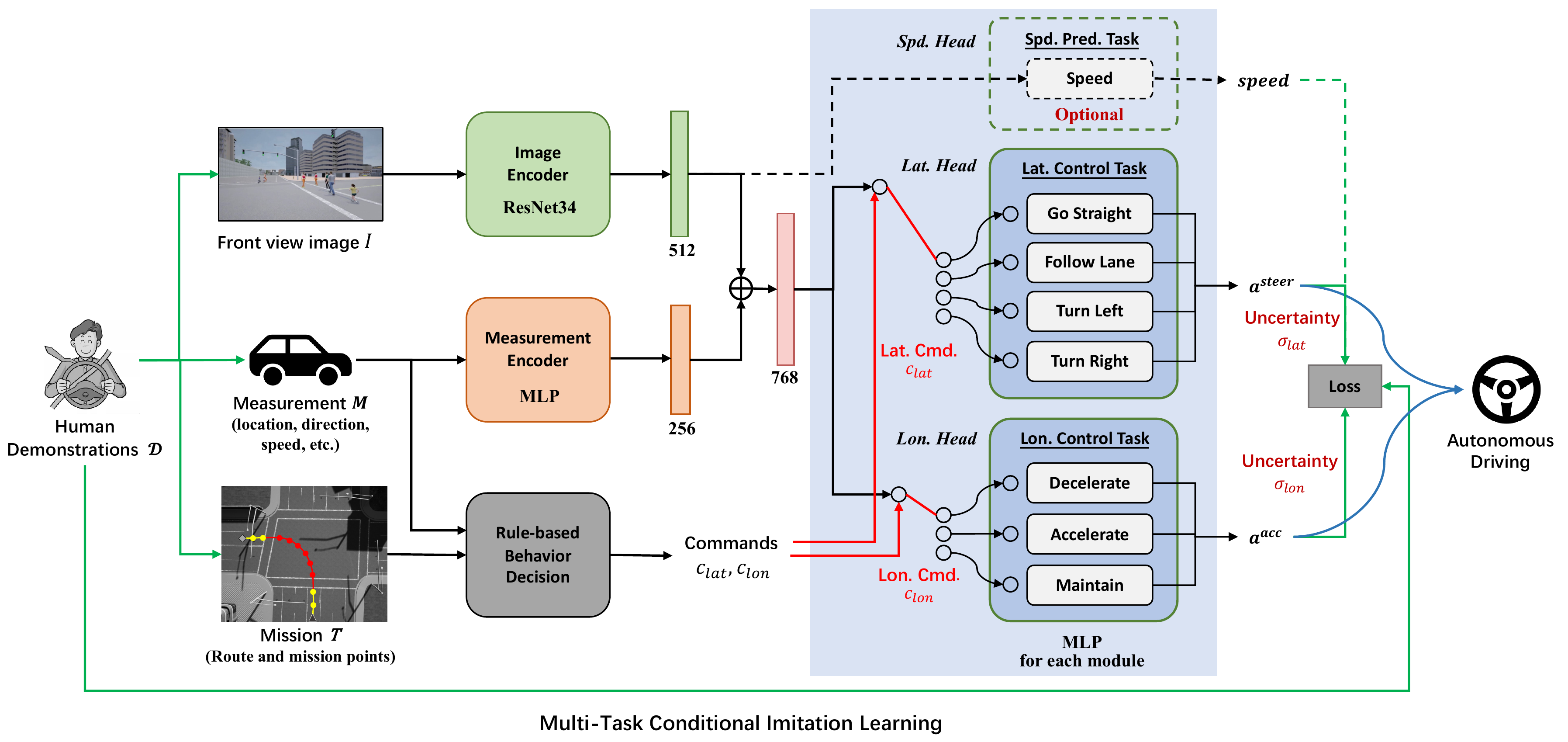}	
\caption{Our proposed multi-task conditional imitation learning (MTCIL) framework, where two separate branches predict lateral and longitudinal control actions, respectively. Both branches share the same perception representation. For each task, corresponding high-level commands are given by rule-based decision module to select the target submodules. Task-dependent uncertainties are learned to adaptively adjust task weights.}
\label{fig:network_architecture}
%\vspace{-5mm}
\end{figure*}
\section{Methodology}
\label{sec:methodology} 
\subsection{Scenario}
This work studies the scenario of an autonomous driving agent navigating through a densely populated intersection, where it needs to adjust its controls and interact safely with pedestrians on crosswalks. In order to have the problem focused, this study does not consider interactions with other vehicles and reactions to traffic signals. The influence of these factors will be further studied in future work.

As shown in Fig. \ref{fig:scene_illustration}, the autonomous vehicle completes the missions of left turn, right turn and go straight at the intersection, guided by the route from a start point to an end point and commands issued by a higher-level module. To accomplish a mission, the agent needs to perform a sequence of driving behaviors, hereinafter referred to as commands, each of which is completed by a sequence of control actions. Specifically, lateral commands include follow lane, go straight, turn left and turn right. Longitudinal commands are decelerate, maintain and accelerate.
\subsection{Conditional Imitation Learning (CIL)}
This research follows the Conditional Imitation Learning \cite{codevilla2018end} framework to formulate the problem as follows: Human driving demonstration dataset ${\cal D}=\{\xi_i\}_{i=1}^{N}$ consist of $N$ trajectories. Each trajectory $\xi_i$ is composed of a sequence of observation-action pairs $\{(o_{i}^t, a_{i}^t, c_{i}^t)\}_{t=1}^{T}$, where $o_{i}^t$, $a_{i}^t$ and $c_{i}^t$ denote the observation, action, and high-level command, respectively. The observations are tuples which include an onboard front-view RGB image $I_{i}^t$ and scalar value ego speed $v_{i}^t$. The actions contain steering angle $a_{i}^{t, str} \in [-1,1]$ and acceleration value $a_{i}^{t, acc} \in [-1, 1]$. The goal is to learn a deep neural network policy $\pi$ parameterized by $\theta$ that imitates human driving behavior. The optimal parameters $\theta^*$ are obtained by minimizing the imitation cost $\cal L$:
\begin{equation}
\theta^* = {\arg min}_{\theta}\sum_j{{\cal L}(\pi(o_j, c_j;\theta),a_j)}
\label{eqn:original_imitation_loss}
\end{equation}

\subsection{Multi-task Learning (MTL)}
Lateral and longitudinal control are two tasks of very different properties. For example, scene features have different importance in accomplishing each task, where lane markings and road structures are more important for lateral control task while obstacles ahead and ego speed have significant influence on the longitudinal control task. Lateral and longitudinal control have different tolerances for vibration in the control actions. Faced with the same scenario, the confidence levels of the lateral and longitudinal controls differ, reflecting the various uncertainties inherent in these tasks.

In multi-task learning, separate deep models are learned for each task and different learning objectives are combined in one loss function \cite{ruder2017overview, zhang2021survey}. Linear combination is typically applied by weighting the losses for each individual task using the hand-tuned hyperparameters \cite{kendall2018multi}. However, the search and tuning of hyperparameters is not trivial. Since model performance is often hyperparameter-sensitive, its versatility may be limited in various scenarios.

Following \cite{kendall2018multi}, this work formulates simultaneous lateral and longitudinal control learning in a multi-task learning framework, where task-dependent uncertainties are used to weight tasks. These uncertainties are also learned from data and optimized simultaneously with model parameters.
\begin{figure*}
\vspace{3mm}
	\centering
	\includegraphics[width=0.8\textwidth]{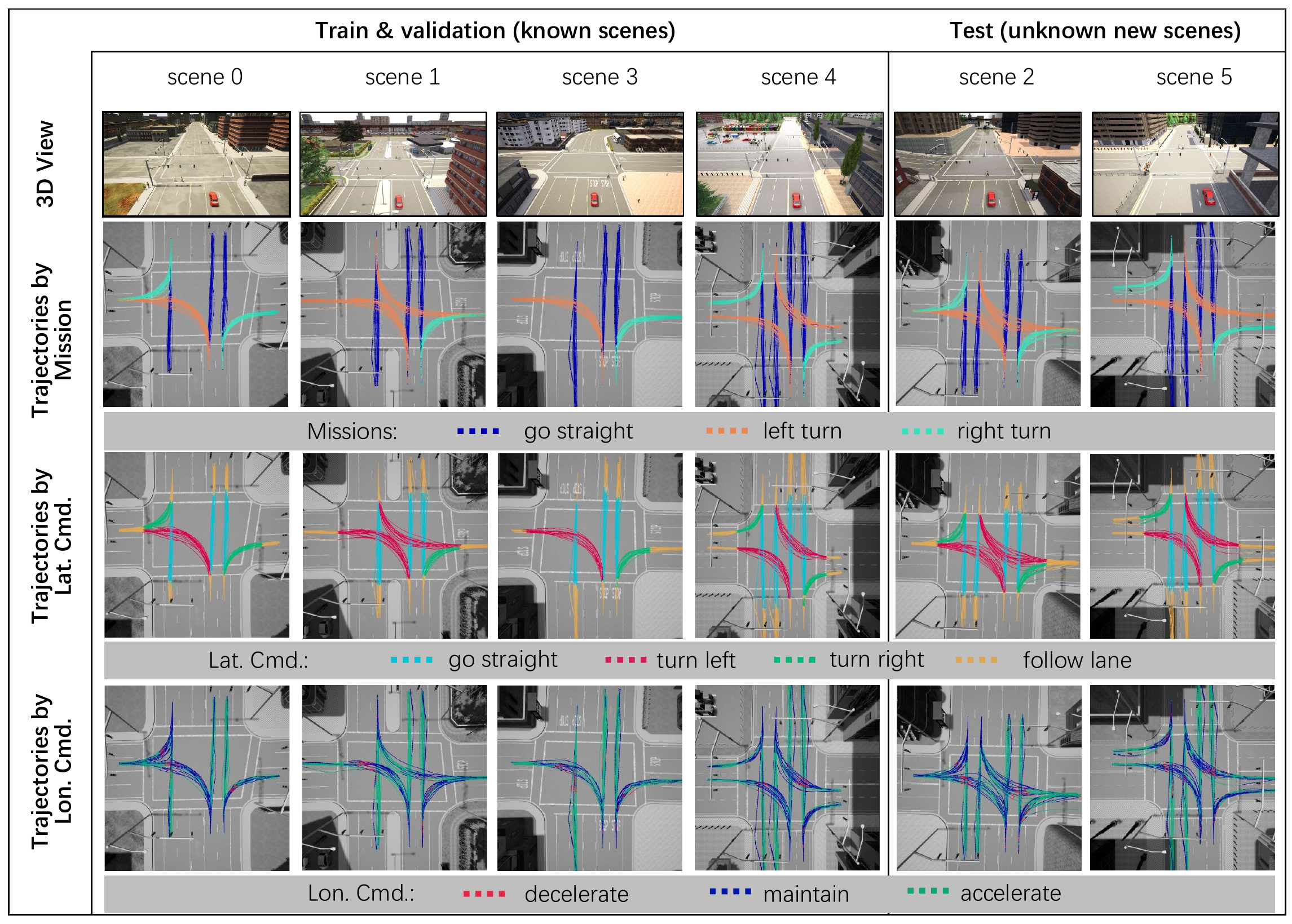}
	\caption{Benchmark scenes and human demonstration trajectories. (Better view in color)}
	\label{fig:benchmark_scenarios} 
%	\vspace{-3mm}
\end{figure*}

\subsection{Task-dependent Uncertainty Loss}
We derive from a single regression task such as learning only lateral or longitudinal control. Let $\pi_\theta(s)$ be a DNN policy model with parameter $\theta$, which takes input data $s$ and outputs control action $a$. The likelihood is modeled as a Gaussian with the mean given by the model output, and the noise scalar $\sigma^2$ represents task-dependent uncertainty:
\begin{eqnarray}
	\displaystyle p(a|\pi_\theta(s)) &=& {\cal N}(\pi_\theta(s), \sigma^2) \\
	\displaystyle  - \log p(a|\pi_\theta(s)) &\propto & \frac{1}{2\sigma^2} \Vert a -\pi_\theta(s) \Vert^2+\log \sigma
\end{eqnarray}
Now consider a multi-task problem that yields two outputs $a_1$ and $a_2$. Assuming the independence of two tasks, we have:
\begin{eqnarray}
	\displaystyle p(a_1,a_2|\pi_\theta(s)) &=& p(a_1|\pi_\theta(s)) \cdot p(a_2|\pi_\theta(s)) \notag \\
	\displaystyle &=& {\cal N}(a_1; \pi_\theta(s), \sigma_1^2) \cdot {\cal N}(a_2; \pi_\theta(s), \sigma_2^2) \notag \\
\end{eqnarray}
\begin{eqnarray}
	\displaystyle - \log p(a_1,a_2|\pi_\theta(s)) &\propto&  \frac{1}{2\sigma_1^2} \Vert a_1 -\pi_\theta(s) \Vert^2 \notag \\
	\displaystyle  &+& \frac{1}{2\sigma_2^2} \Vert a_2 -\pi_\theta(s) \Vert^2+\log \sigma_1\sigma_2 \notag \\
	\label{eqn:mtl_loss}
\end{eqnarray}
Consequently, we have the task-dependent uncertainty loss for the multi-task learning of lateral and longitudinal controls:
\begin{eqnarray}
		\displaystyle {\cal L} (\theta, \sigma_{lat}, \sigma_{lon}) &=& \frac{1}{2\sigma_{lat}^2} \Vert a^{str} -\pi_\theta^{str}(s) \Vert^2  \notag \\
		\displaystyle &+& \frac{1}{2\sigma_{lon}^2} \Vert a_{acc} -\pi_\theta^{acc}(s) \Vert^2+\log \sigma_{lat}\sigma_{lon} \notag
\end{eqnarray}
where $\pi_\theta(s)$ denotes $\pi(o,c;\theta)$, which is composed of three sub DNN models, i.e., a feature encoder $\pi_\theta^{feat}$ shared by the lateral and longitudinal conditional modules $\pi_\theta^{str}$ and $\pi_\theta^{acc}$. $\sigma_{lat}$ and $\sigma_{lon}$ denote the task-dependent uncertainties of lateral and longitudinal controls, respectively. We can interpret the first and second terms in the loss function as the objectives of each individual task, which are weighted by $\sigma_{lat}$ and $\sigma_{lon}$, respectively. Minimizing the loss function with respect to $\sigma_{lat}$ and $\sigma_{lon}$ can learn their relative weights from data. For example, large $\sigma_{lat}$ implies that the lateral control task is inherently more uncertain, then we have a smaller weight of the task, and vice versa. Different from literature where the weights of steering and acceleration losses are manually tuned hyperparameters, our method adaptively learns to balance between them. The last term $\log \sigma_{lat}\sigma_{lon}$ serves as a regularization for preventing $\sigma_{lat}$ and $\sigma_{lon}$ from increasing too much.

\subsection{Multi-Task Conditional Imitation Learning}
The proposed Multi-Task Conditional Imitation Learning (MTCIL) architecture is shown in Fig. \ref{fig:network_architecture}. We take the single-frame front view image $I$ and the ego velocity value $v$ as the input to the image encoder and measurement encoder, respectively. For image encoders, we evaluate the performance of CarlaNet \cite{codevilla2018end} and ResNet34 \cite{he2016deep} in the experiments. The measurement encoder is a multi-layer perceptron (MLP) consisting of three fully connected layers. The concatenated features from two encoders are passed to the control modules. The lateral and longitudinal control tasks are completed by a conditional module, which contains multiple MLPs corresponding to each lateral or longitudinal command. Given current commands $C_{lat}$ and $C_{lon}$ determined by a rule-based model, the corresponding modules are switched on and responsible for predicting control actions $a^{str}$ and $a^{acc}$.
% \usepackage[normalem]{ulem}
\begin{table*}[]
\vspace{3mm}
\centering
\caption{Considered events in our benchmark and comparison to other benchmarks}
\label{tab:benchmark_comparison}
\begin{tabular}{clll}
\hline
\textbf{Benchmark} &
  \multicolumn{1}{c}{\textbf{Failure events}} &
  \multicolumn{1}{c}{\textbf{Definition of success}} &
  \multicolumn{1}{c}{\textbf{Metrics}} \\ \hline
\textbf{\begin{tabular}[c]{@{}c@{}}original \\ CARLA\\ benchmark \cite{dosovitskiy2017carla}\end{tabular}} &
  \begin{tabular}[c]{@{}l@{}}1. collision with static object/car/pedestrian\\ 2. opposite lane\\ 3. sidewalk\end{tabular} &
  \begin{tabular}[c]{@{}l@{}}The agent reaches the goal \textbf{regardless of} \\ what happened during the episode.\end{tabular} &
  \begin{tabular}[c]{@{}l@{}}1. success rate\\ 2. avg. distance travelled\\ between infractions\end{tabular} \\ \hline
\textbf{\begin{tabular}[c]{@{}c@{}}NoCrash\\ benchmark \cite{codevilla2019exploring}\end{tabular}} &
  \begin{tabular}[c]{@{}l@{}}1. collision with static object/car/pedestrian\\ 2. timeout\\ 3. traffic light violations\end{tabular} &
  \begin{tabular}[c]{@{}l@{}}The agent reaches the goal under a time \\ limit without colliding with any object.\end{tabular} &
  \begin{tabular}[c]{@{}l@{}}1. success rate\\ 2. collision rate\\ 3. timeout rate\end{tabular} \\ \hline
\textbf{\begin{tabular}[c]{@{}c@{}}CARLA\\ Leaderboard \cite{carlaleaderboard}\end{tabular}} &
  \begin{tabular}[c]{@{}l@{}}1. collision with static object/car/pedestrian\\ 2. running a red light/stop sign\\ 3. timeout\end{tabular} &
  not applicable &
  \begin{tabular}[c]{@{}l@{}}1. driving score\\ 2. route completion rate\\ 3. infraction penalty\end{tabular} \\ \hline
\textbf{\begin{tabular}[c]{@{}c@{}}IntersectNav\\ benchmark\\ (ours)\end{tabular}} &
  \begin{tabular}[c]{@{}l@{}}1. \textbf{collision} with static object/car/pedestrian\\ 2. \textbf{lane invasion} (invasion time $>$ 5)\\ 3. \textbf{poor end pose} (The agent approaches\\the end point, but its heading's deviation\\from lane direction $>$ 15\degree or vertical \\deviation from lane centerline $>$ 1m)\\ 4. \textbf{timeout} (failure to arrive at the goal \\within 1000 steps)\end{tabular} &
  \begin{tabular}[c]{@{}l@{}}The agent reaches the goal under a time \\ limit \textbf{without any failure events happened}\end{tabular} &
  \begin{tabular}[c]{@{}l@{}}1. success rate\\ 2. collision rate\\ 3. timeout rate\\ 4. lane invasion rate\\ 5. poor end pose rate\\ 6. other metrics reflecting\\ control quality (see Tab.\ref{tab:metrics_description})\end{tabular} \\ \hline
\end{tabular}
%\vspace{-4mm}
\end{table*}
\begin{table*}
\vspace{3mm}
\centering
\caption{Metrics that reflect the control quality}
\label{tab:metrics_description}
\resizebox{0.95\linewidth}{!}{%
\begin{tabular}{ccc}
\hline
\textbf{Metric(Unit)} &
  \textbf{Description} &
  \multicolumn{1}{c}{\textbf{Formula}} \\ \hline
\textbf{Ego Jerk(\#)} &
  \begin{tabular}[c]{@{}c@{}}Average times of the absolute values of control actions \textgreater 0.9\end{tabular} &
  $\frac{1}{N}\sum_{i=1}^{N}\sum_{t=1}^{T_i}\mathds {1}[|a^{str}_{t}|>0.9\ or\ |a^{acc}_{t}|>0.9]$ \\ \hline
\textbf{Other Jerk(\#)} &
  \begin{tabular}[c]{@{}c@{}}Average times of pedestrians $p_j, j=1...M_i$, disrupted by\\ ego agent (e.g., emergent stop in close range)\end{tabular} &
  $\frac{1}{N}\sum_{i=1}^{N}\sum_{t=1}^{T_i}\sum_{j=1}^{M_i}\mathds {1}[p_j.get\_disrupted()=True]$ \\ \hline
\begin{tabular}[c]{@{}c@{}}\textbf{Deviation from}\\ \textbf{Waypoint(m)}\end{tabular} &
  \begin{tabular}[c]{@{}c@{}}Mean location $\overrightarrow{loc}_t$'s deviation from centerline represented\\ by the current nearest waypoint $\overrightarrow{{wp}}_{t}^{c}$ and next waypoint $\overrightarrow{{wp}}_{t}^{n}$ \end{tabular} &
  $\frac{1}{N}\sum_{i=1}^{N}\sum_{t=1}^{T_i} \frac{(\overrightarrow{{wp}}_{t}^{n} - \overrightarrow{{wp}}_{t}^{c}) \times (\overrightarrow{loc}_t - \overrightarrow{{wp}}_{t}^{c})}{|\overrightarrow{{wp}}_{t}^{n} - \overrightarrow{{wp}}_{t}^{c}|}$ \\ \hline
\begin{tabular}[c]{@{}c@{}}\textbf{Deviation from} \\ \textbf{Destination(m)}\end{tabular} &
  \begin{tabular}[c]{@{}c@{}}Mean final location $\overrightarrow{loc}_{T_i}$'s deviation from the goal location $\overrightarrow{g}_i$\end{tabular} &
  $\frac{1}{N}\sum_{i=1}^{N} |\overrightarrow{loc}_{T_i} - \overrightarrow{g}_i|$ \\ \hline
\begin{tabular}[c]{@{}c@{}}\textbf{Heading Angle}\\ \textbf{Deviation(\degree)}\end{tabular} &
  \begin{tabular}[c]{@{}c@{}}Mean final heading $\theta_{T_i}$'s deviation from lane direction $\delta_i$ \\at the episode ending\end{tabular} &
  $\frac{1}{N}\sum_{i=1}^{N} |\theta_{T_i} - \delta_i|$ \\ \hline
\textbf{Total Step(\#)} &
  Average total steps for each episode &
  $\frac{1}{N}\sum_{i=1}^{N} T_i$ \\ \hline
\end{tabular}
}
%\vspace{-5mm}
\end{table*}

Compared with literature work that uses a single deep model to output both lateral and longitudinal control actions, separate modeling can greatly improve the performance of longitudinal control, which is crucial for dense intersections with pedestrian interactions, as shown in experiments. Furthermore, combining both controls into a multi-task framework can improve efficiency by sharing encoders, while balancing performance by weighting both tasks according to task-dependent uncertainties, which can be learned automatically from data. Note that this framework can be easily extended to allow more tasks such as speed prediction. Similar to \cite{codevilla2019exploring}, an optional branch predicting the current speed can be added in our framework (see Fig. \ref{fig:network_architecture}), which encourages the perception module to extract visual cues that reflect the scene dynamics. The performance is examined in the experiments.
\section{A New Benchmark: IntersectNav}
\label{sec:benchmark}
We propose a new benchmark named IntersectNav in this section. Unlike the CoRL2017 benchmark \cite{dosovitskiy2017carla} and the NoCrash benchmark \cite{codevilla2019exploring}, we focus on intersections that challenge and extensively analyze the ability of driving agents to interact with pedestrians. Specifically, we use CARLA \cite{dosovitskiy2017carla} driving simulator 0.9.7 for realistic 3D simulation. Compared to the 0.8.X version used in previous benchmarks \cite{dosovitskiy2017carla, codevilla2019exploring}, the graphics and simulation behavior changed a lot in 0.9.7, making it more complex and realistic. 
\subsection{Scenarios}
Demonstrated in Fig. \ref{fig:benchmark_scenarios}, six different US-style unsignalized intersections from two towns are selected for evaluation. Four scenes are used for train and validation while the other two are reserved for test. We configure the available start and goal points, which define the reference routes (adds up to 40). The benchmark adopts an episodic setup. At each episode, an intersection is chosen and the ego car randomly starts from one of the available configurations. The world weather is randomly selected from \{ClearNoon, CloudyNoon, WetNoon, HardRainNoon\}. Other four new weathers \{ClearSunset, CloudySunset, WetSunset, HardRainSunet\} are reserved for test. Three missions are considered, i.e., performing left turn/go straight/right turn and navigate through the intersection (c.f. Fig. \ref{fig:benchmark_scenarios} row 2). A random number of 20-30 pedestrians are generated to walk through the crosswalks  around intersections. Our setup ensures that the driving agent will inevitably encounter pedestrians during the course of turning. Although only pedestrians are considered in current settings, our benchmark can be easily extended to consider other vehicles, traffic lights and signs.

\subsection{Evaluation}
\label{sec:evaluation_metrics}
During the close-loop simulation for evaluation, the ego agent and pedestrians are initialized according to protocols described above. At each simulation step, current observations and commands are fed into the control model. The network's control outputs (both lateral and longitudinal) are then clipped by the range $[-1.0, 1.0]$ and passed to the actuators in CARLA. The backend engine simulates the world dynamics and moves on to the next step. This process iterates until an episode is terminated. We consider five possible events that the episode ends with: collision, lane invasion, poor end pose, timeout and success. Detailed information can be found in Tab. \ref{tab:benchmark_comparison}, which also compares with other benchmarks. Note that our benchmark sets up higher requirements of the model's control precision through lane invasion and poor end pose metrics.

Aside from above metrics that consider task completeness, we also define metrics to evaluate the model's control quality. The details are provided in Tab. \ref{tab:metrics_description}. By introducing the statistics of ego and other jerks, we can further analysis the ego's driving comfort along with its influence on other pedestrians. The deviations consider the control precision while total steps measure how efficient is the learned model.
\subsection{Human Demonstration Dataset}
\begin{figure}[h]
	\centering
	\includegraphics[width=0.9\linewidth]{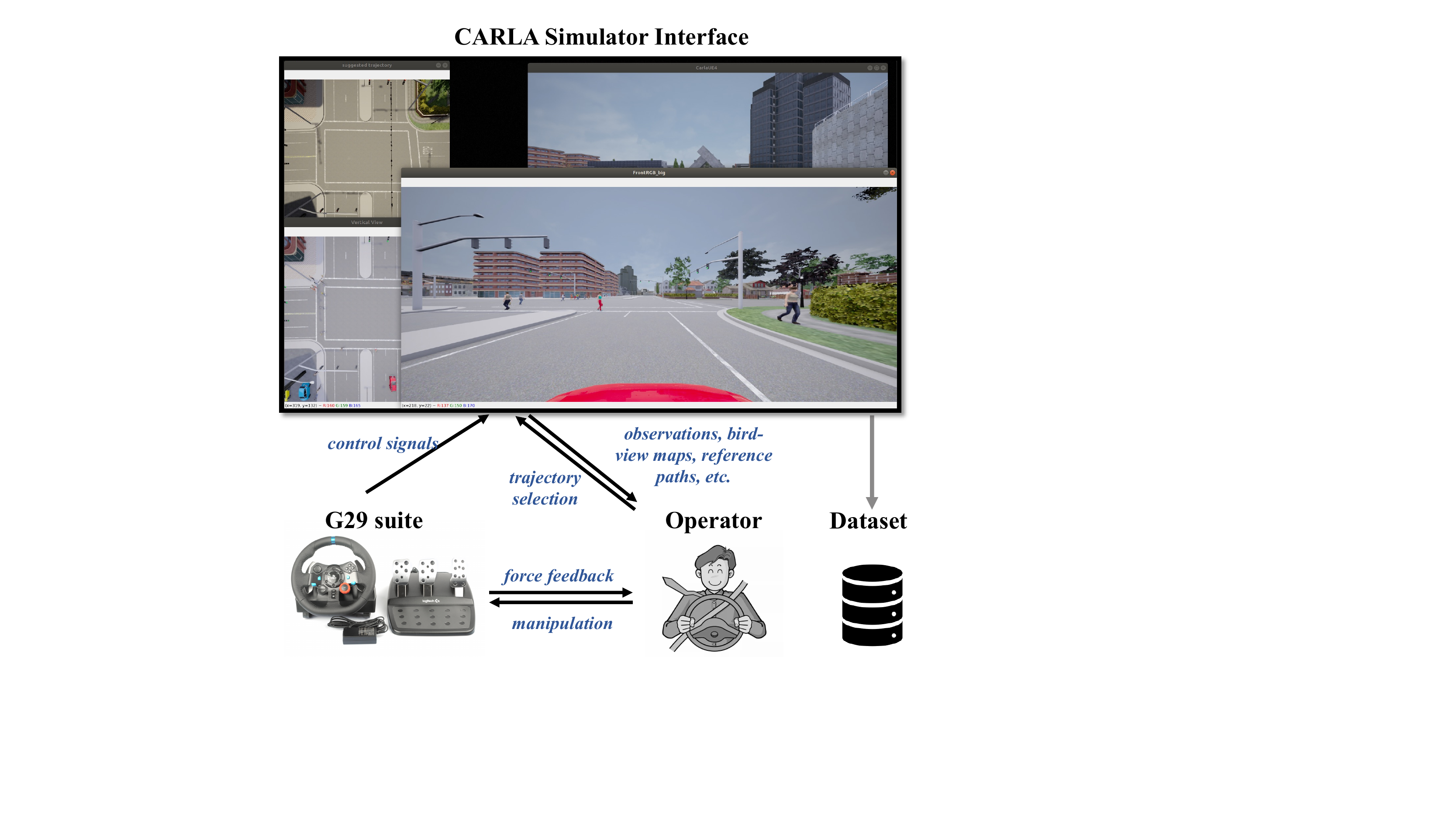}
	\caption{Data collection procedure. The human operator manipulates the driving suite to demonstrate the mission in CARLA simulator.}
	\label{fig:data_collect}
	\vspace{-3mm}
\end{figure}

\begin{figure}[h]
	\centering
	\includegraphics[width=0.85\linewidth]{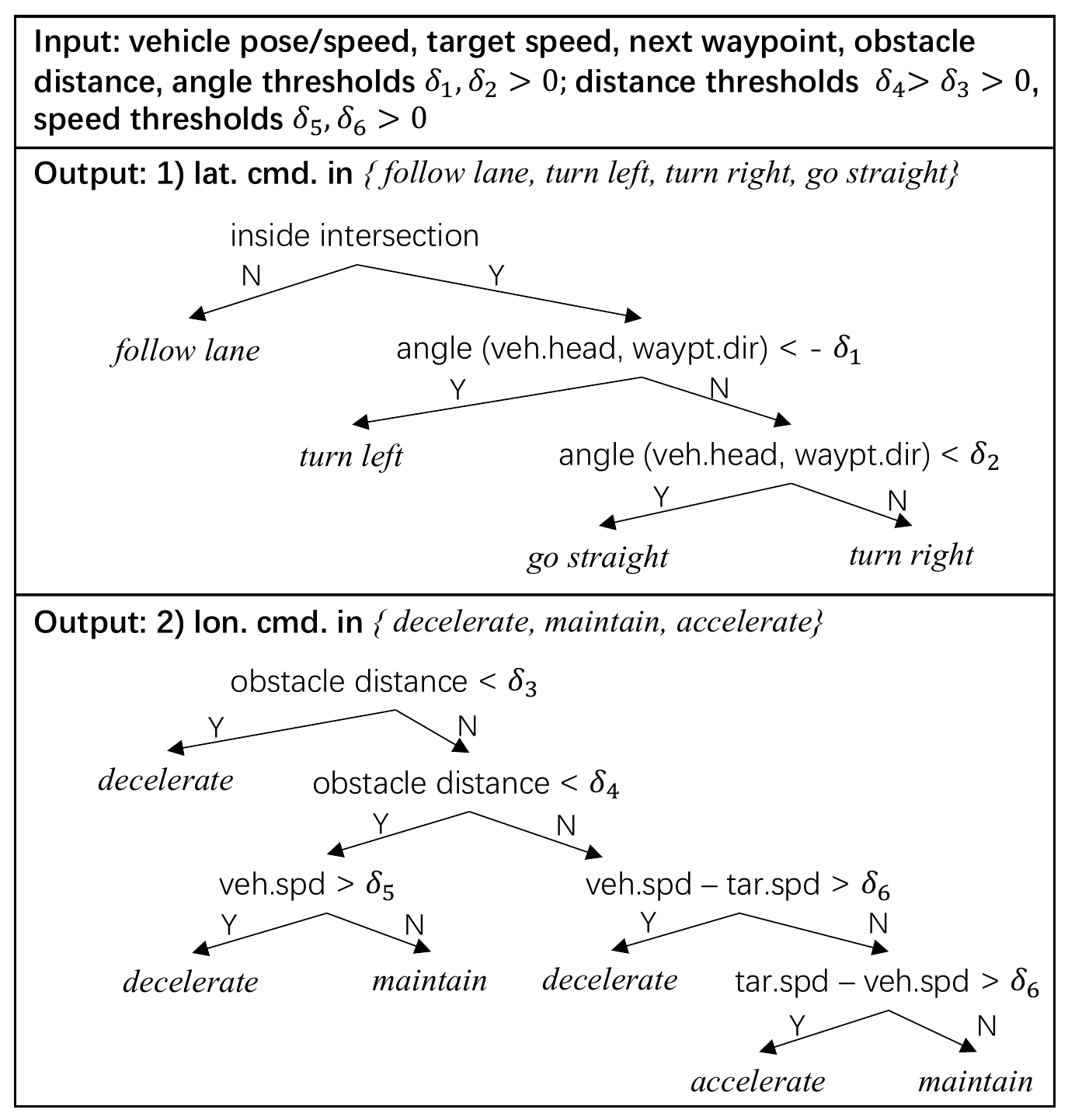}
	\caption{Rule-based decision module.}
	\label{fig:rule_based_decision}
	\vspace{-4mm}
\end{figure}

As is shown in Fig. \ref{fig:data_collect}, we collect human driving demonstrations in CARLA through the driving suite that includes a dual-motor force feedback wheel and a floor pedal. The human driver is provided with real-time front-view RGB images and bird-view images. Reference routes are projected onto the bird-view map to provide the mission information. Real-time high-level driving commands from a rule-based decision module (cf. Fig. \ref{fig:rule_based_decision}) are provided for reference. In each episode, the operator is asked to keep a preferred 20 km/h speed and drive through the intersection following the high-level commands. 
\begin{table}[h]
\vspace{3mm}
\centering
\caption{Statistics of the human demonstration dataset}
\label{tab:benchmark_statistics}
\resizebox{0.8\linewidth}{!}{%
\begin{tabular}{clccclclcccl}
\hline
\multicolumn{12}{c}{\textbf{Frames (Trajectories) by Scene}} \\ \hline
\multicolumn{2}{c|}{Scene 0} & \multicolumn{2}{c|}{Scene 1} & \multicolumn{2}{c|}{Scene 3} & \multicolumn{2}{c|}{Scene 4} & \multicolumn{2}{c|}{Scene 2} & \multicolumn{2}{c}{Scene 5} \\
\multicolumn{2}{c|}{\begin{tabular}[c]{@{}c@{}}12113 \\ (150)\end{tabular}} & \multicolumn{2}{c|}{\begin{tabular}[c]{@{}c@{}}10589 \\ (147)\end{tabular}} & \multicolumn{2}{c|}{\begin{tabular}[c]{@{}c@{}}8312 \\ (119)\end{tabular}} & \multicolumn{2}{c|}{\begin{tabular}[c]{@{}c@{}}13350 \\ (186)\end{tabular}} & \multicolumn{2}{c|}{\begin{tabular}[c]{@{}c@{}}16737 \\ (199)\end{tabular}} & \multicolumn{2}{c}{\begin{tabular}[c]{@{}c@{}}14682 \\ (187)\end{tabular}} \\ \hline
\multicolumn{12}{c}{\textbf{Frames (Trajectories) by Mission}} \\ \hline
\multicolumn{4}{c|}{Left turn} & \multicolumn{4}{c|}{Go straight} & \multicolumn{4}{c}{Right turn} \\
\multicolumn{4}{c|}{25952 (229)} & \multicolumn{4}{c|}{25253 (515)} & \multicolumn{4}{c}{24578 (244)} \\ \hline
\multicolumn{12}{c}{\textbf{Frames by Lat. Cmd.}} \\ \hline
\multicolumn{3}{c|}{Follow lane} & \multicolumn{3}{c|}{Turn   left} & \multicolumn{3}{c|}{Turn  right} & \multicolumn{3}{c}{Go straight} \\
\multicolumn{3}{c|}{34812} & \multicolumn{3}{c|}{18015} & \multicolumn{3}{c|}{13719} & \multicolumn{3}{c}{9237} \\ \hline
\multicolumn{12}{c}{\textbf{Frames by Lon. Cmd.}} \\ \hline
\multicolumn{4}{c|}{Decelerate} & \multicolumn{4}{c|}{Maintain} & \multicolumn{4}{c}{Accelerate} \\
\multicolumn{4}{c|}{16258} & \multicolumn{4}{c|}{25432} & \multicolumn{4}{c}{34093} \\ \hline
\end{tabular}
}
%\vspace{-4mm}
\end{table}
 
At each time step, a small random uniform noise is added to the human's steering with probability 0.1. This technique aims to collect experts' demonstrations that recover from perturbations. Once an episode is over, the operator can review this episode's metrics in Tab. \ref{tab:metrics_description}. Data from successful episodes with good control metrics is stored. We record raw sensor data (e.g., RGB/depth images, ego's speed and poses etc.) along with the expert's demonstrations (e.g., control steer/throttle/brake, corresponding high-level commands). The observation $o^t$, expert action $a^t$ and high-level commands $c^t=(c_{lat}^t, c_{lon}^t)$ are bounded together as one tuple $(o^t,a^t,c^t)$, which serves as a training sample. Meta task information such as town/scene/pose index and weather are also recorded, .

We collected over 30 hours of human driving data at six intersections, which contains more than 800 trajectories. Illustration of collected human trajectories is provided in Fig. \ref{fig:benchmark_scenarios}, where the colors represent the different missions/lateral commands/longitudinal commands in \engordnumber{2}/\engordnumber{3}/\engordnumber{4} rows, respectively. Detailed statistics on the number of samples and trajectories can be found in Tab. \ref{tab:benchmark_statistics}. The dataset covers four training weathers, where the proportion of ClearNoon : CloudyNoon : WetNoon : HardRainNoon is about 0.45 : 0.17 : 0.18 : 0.19. The data from four intersections is split into the train dataset and validation dataset at a ratio of approximately 5:1. Data from the other two intersections is used for test.

\begin{table}[h]
\vspace{3mm}
\centering
\caption{Evaluation results of task completeness. Abbreviations: success rate (\textbf{SR}), poor end pose rate (\textbf{PR}), timeout rate (\textbf{TR}), lane invasion rate (\textbf{LR}), collision rate (\textbf{CR}).}
\label{tab:comparison_baselines_task_completeness}
\begin{tabular}{ccccccc}
\hline
\textbf{Condition} &
  \textbf{Model} &
  \textbf{\begin{tabular}[c]{@{}c@{}}SR $\uparrow$\end{tabular}} &
  \textbf{\begin{tabular}[c]{@{}c@{}}PR $\downarrow$\end{tabular}} &
  \textbf{\begin{tabular}[c]{@{}c@{}}TR $\downarrow$\end{tabular}} &
  \textbf{\begin{tabular}[c]{@{}c@{}}LR $\downarrow$\end{tabular}} &
  \textbf{\begin{tabular}[c]{@{}c@{}}CR $\downarrow$\end{tabular}} \\ \hline
\multirow{3}{*}{\textbf{\begin{tabular}[c]{@{}c@{}}Train scene\\ \&\\ Train weather\end{tabular}}} &
  CIL &
  59.5 &
  8.7 &
  29.4 &
  0.8 &
  1.6 \\
 &
  CILRS &
  63.2 &
  3.2 &
  4.8 &
  10.4 &
  18.4 \\
 &
  Ours &
  \textbf{87.6} &
  7.6 &
  \textbf{2.4} &
  \textbf{0.0} &
  2.4 \\ \hline
\multirow{3}{*}{\textbf{\begin{tabular}[c]{@{}c@{}}Test scene \\ \&\\ Train weather\end{tabular}}} &
  CIL &
  66.3 &
  0.0 &
  27.5 &
  2.5 &
  3.8 \\
 &
  CILRS &
  57.5 &
  0.0 &
  17.5 &
  6.2 &
  18.8 \\
 &
  Ours &
  \textbf{92.5} &
  2.5 &
  \textbf{0.0} &
  \textbf{2.5} &
  \textbf{2.5} \\ \hline
\multirow{3}{*}{\textbf{\begin{tabular}[c]{@{}c@{}}Test scene\\ \&\\ Test weather\end{tabular}}} &
  CIL &
  51.2 &
  0 &
  40.0 &
  1.3 &
  7.5 \\
 &
  CILRS &
  43.2 &
  4.0 &
  40.0 &
  3.2 &
  9.6 \\
 &
  Ours &
  \textbf{82.5} &
  3.8 &
  \textbf{1.2} &
  12.5 &
  \textbf{0.0} \\ \hline
\end{tabular}
%\vspace{-4mm}
\end{table}
% \usepackage{multirow}
\begin{table*}[h]
\vspace{3mm}
\centering
\caption{Evaluation results of control quality}
\resizebox{0.85\linewidth}{!}{%
\begin{tabular}{cccccccc}
\hline
\textbf{Condition} &
  \textbf{Model} &
  \textbf{\begin{tabular}[c]{@{}c@{}}Ego Jerk\\ \#, $\downarrow$\end{tabular}} &
  \textbf{\begin{tabular}[c]{@{}c@{}}Other Jerk\\ \#, $\downarrow$\end{tabular}} &
  \textbf{\begin{tabular}[c]{@{}c@{}}Deviation from \\Waypoint m, $\downarrow$\end{tabular}} &
  \textbf{\begin{tabular}[c]{@{}c@{}}Deviation from \\Destination m, $\downarrow$\end{tabular}} &
  \textbf{\begin{tabular}[c]{@{}c@{}}Heading Angle\\Deviation \degree, $\downarrow$\end{tabular}} &
  \textbf{\begin{tabular}[c]{@{}c@{}}Total Steps\\ \#, $\downarrow$\end{tabular}} \\ \hline
\multirow{3}{*}{\textbf{\begin{tabular}[c]{@{}c@{}}Train scene\\ \&\\ Train weather\end{tabular}}}  & CIL   & 0.294      & 43.96           & 1.4   & 5.248          & 10.618          & 488.556          \\
                                & CILRS & 0          & 20.872          & 0.429 & 3.988          & 11.122          & 226.456          \\
                                & Ours  & \textbf{0} & 55.088          & 0.658 & \textbf{1.588} & 5.472 & 326.008          \\ \hline
\multirow{3}{*}{\textbf{\begin{tabular}[c]{@{}c@{}}Test scene\\ \&\\ Train weather\end{tabular}}}   & CIL   & 0.125      & 68.988          & 1.286 & 5.286          & 9.83          & 505.7            \\
                                & CILRS & 0          & 173.725         & 0.545 & 6.767          & 12.762           & 376.062          \\
                                & Ours  & \textbf{0} & \textbf{12.863} & 0.606 & \textbf{1.153} & 4.182 & \textbf{318.375} \\ \hline
                        
\multirow{3}{*}{\textbf{\begin{tabular}[c]{@{}c@{}}Test scene\\ \&\\ Test weather\end{tabular}}} &
  CIL &
  0.038 &
  67.7 &
  0.938 &
  8.713 &
  17.185 &
  537.888 \\
 &
  CILRS &
  0 &
  106.648 &
  0.494 &
  10.685 &
  19.992 &
  539.384
  \\
 &
  Ours &
  \textbf{0} &
  \textbf{31.438} &
  0.627 &
  \textbf{1.038} &
  3.927 &
  \textbf{333.55}
  \\ \hline
\end{tabular}
}
\label{tab:comparison_baselines_control_quality}
\end{table*}
\begin{figure*}[ht]
	\centering
	\includegraphics[width=0.8\linewidth]{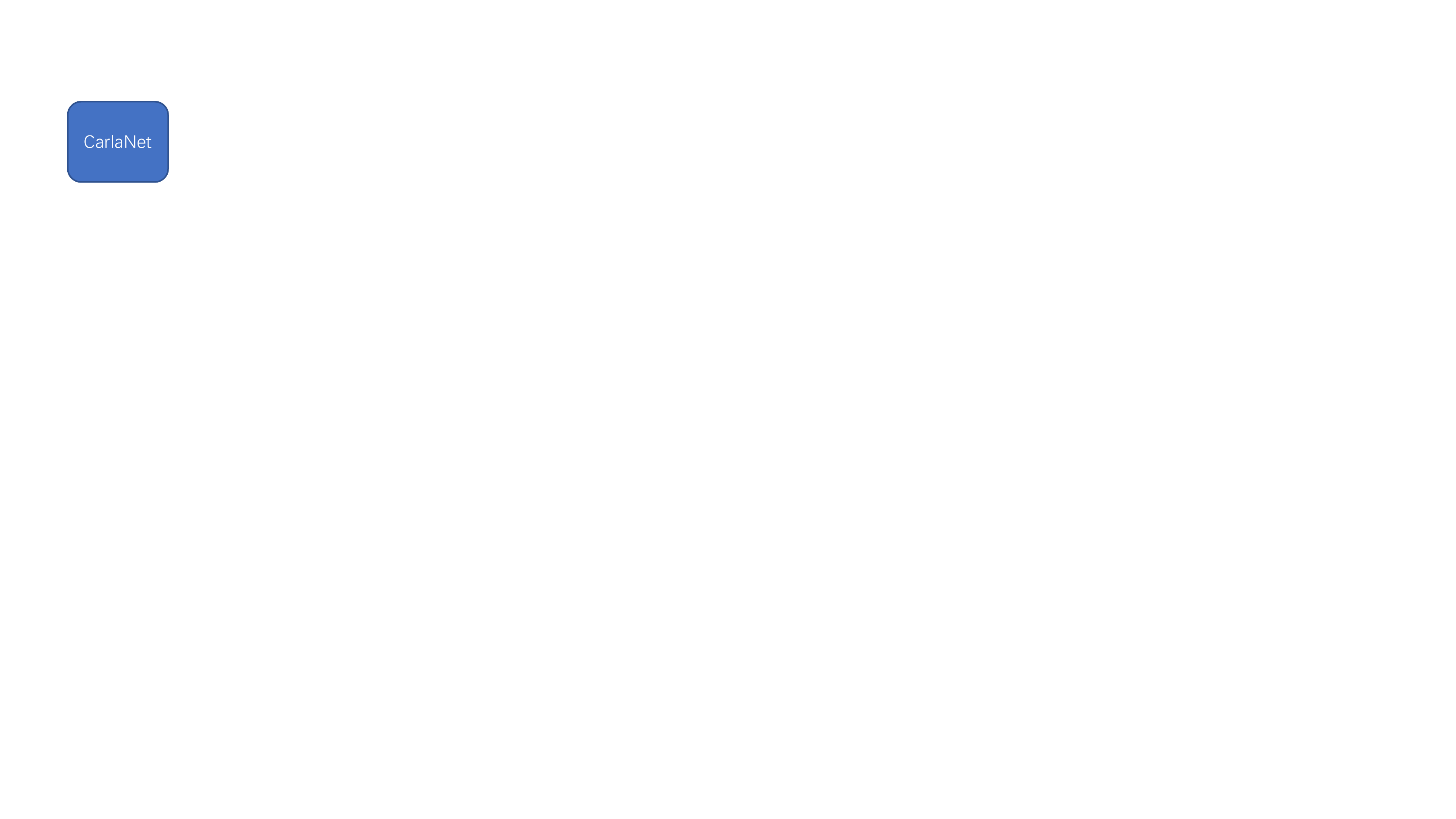}
	\caption{Models for comparison in ablation studies.}
	\label{fig:ablation_study_design}
	\vspace{-6mm}
\end{figure*}
\section{Experiment}
\label{sec:experiment}
\subsection{Training Details}
All models are trained using Adam optimizer \cite{kingma2014adam} with an initial learning rate 2e-4, which will be divided by 10 if validation loss stops decreasing for more than 5 epochs. Dropout is used after fully-connected layers with a probability of 0.5. Each minibatch contains 120 samples, which are randomly sampled from the shuffled trainset. We follow Codevilla et.al and employ a 200 $\times$ 88 image resolution for CarlaNet \cite{codevilla2018end} perception backbone. For ResNet34 backbone, we resize the image to resolution 224 $\times$ 224. If specified, online image augmentation is performed during training, which includes Gaussian blur and noise, dropout, adjust of brightness, contrast and etc. Our results demonstrate the effectiveness of data augmentation, especially for ResNet34 backbone.

\subsection{Evaluation Results}
Since offline and online methods cannot be directly compared, this work focuses on offline methods and chooses CIL \cite{codevilla2018end} and CILRS \cite{codevilla2019exploring} as our baselines where no additional supervisions (e.g., reconstructions, BEV representations) apart from expert demonstrations are used. Multiple episodes for each route in our benchmark are simulated to calculate the average metrics. The evaluation results of task completeness are presented in Tab. \ref{tab:comparison_baselines_task_completeness}. Our reported multi-task model uses ResNet34 backbone and uncertainty loss. 

CIL and CILRS have a similar success rate near 60\% on train and test scenes with train weather. When facing new weathers, the success rates of both models decrease much. Besides, CIL suffers from a large timeout rate ($\sim$30\%). We regard this as the inertia problem \cite{codevilla2019exploring}, where the model creates a spurious correlation between low speed and no acceleration, inducing excessive stopping and difﬁcult restarting. CILRS mitigates the problem by introducing the speed prediction branch. However, CILRS suffers from higher collision rate. These failures show that baselines have difficulty in learning longitudinal control under interactive scenarios. 
Compared to baselines, our method achieves a 30\% success rate gain, which demonstrates the effectiveness of multi-task learning.

Evaluation results of control quality are provided in Tab. \ref{tab:comparison_baselines_control_quality}, which demonstrate that our method has a better control quality than baselines in most conditions.
Our model achieves smaller ego jerk and other jerk, which means more comfortable driving and less influence on pedestrians. As for deviations, our method achieves the best destination deviation, which is consistent with its highest success rate. It also has a much smaller average total steps than baselines in test conditions, which means higher efficiency. When tested on new conditions, our model shows good generalization ability while baseline models exhibit a large decline in performance.

% \usepackage{multirow}
\begin{table}[]
\vspace{3mm}
\caption{Task completeness evaluation results of ablation studies on \textbf{test scene and test weather}. $\backslash$ means \textbf{without}.}
\label{tab:ablation_task_test}
\resizebox{\linewidth}{!}{%
\begin{tabular}{ccccccc}
\hline
\textbf{Group} &
  \textbf{Model} &
  \textbf{\begin{tabular}[c]{@{}c@{}}SR $\uparrow$\end{tabular}} &
  \textbf{\begin{tabular}[c]{@{}c@{}}PR $\downarrow$\end{tabular}} &
  \textbf{\begin{tabular}[c]{@{}c@{}}TR $\downarrow$\end{tabular}} &
  \textbf{\begin{tabular}[c]{@{}c@{}}LR $\downarrow$\end{tabular}} &
  \textbf{\begin{tabular}[c]{@{}c@{}}CR $\downarrow$\end{tabular}} \\ \hline
\multirow{4}{*}{\textbf{\begin{tabular}[c]{@{}c@{}}Base-\\lines \end{tabular}}} &
  CIL$\backslash$\textbf{Aug} &
  35 &
  1.3 &
  62.5 &
  0.0 &
  1.2 \\
 &
  CILRS$\backslash$\textbf{Aug} &
  8.7 &
  0.0 &
  80.0 &
  2.5 &
  8.8 \\
 &
  CIL &
  51.2 &
  0.0 &
  40.0 &
  1.3 &
  7.5 \\
 &
  CILRS &
  43.2 &
  4.0 &
  40.0 &
  3.2 &
  9.6 \\ \hline
\multirow{4}{*}{\textbf{\begin{tabular}[c]{@{}c@{}}MT+\\ hLoss\end{tabular}}} &
  CN+\textbf{MT}+\textbf{hLoss}$\backslash$\textbf{Aug} &
  50.0 &
  18.8 &
  1.2 &
  20.0 &
  10.0 \\
 &
  RN+\textbf{MT}+\textbf{hLoss}$\backslash$\textbf{Aug} &
  71.2 &
  2.5 &
  5.0 &
  3.8 &
  17.5 \\
 &
  CN+\textbf{MT}+\textbf{hLoss} &
  72.5 &
  5.0 &
  2.5 &
  1.2 &
  18.8 \\
 &
  RN+\textbf{MT}+\textbf{hLoss} &
  67.5 &
  5.0 &
  1.3 &
  5.0 &
  21.2 \\ \hline
\multirow{2}{*}{\textbf{\begin{tabular}[c]{@{}c@{}}MT+\\ uLoss \end{tabular}}} &
  \begin{tabular}[c]{@{}c@{}}CN+\textbf{MT}+\textbf{uLoss}\textbf{(Ours)}\end{tabular} &
  81.2 &
  5.0 &
  1.3 &
  11.2 &
  1.3 \\
 &
  \begin{tabular}[c]{@{}c@{}}RN+\textbf{MT}+\textbf{uLoss}\textbf{(Ours)}\end{tabular} &
  82.5 &
  3.8 &
  1.2 &
  12.5 &
  0.0 \\ \hline
\end{tabular}
}
\vspace{-6mm}
\end{table}

\subsection{Ablation Studies}
Ablation experiments in Fig. \ref{fig:ablation_study_design} are conducted to further investigate the importance of three components: backbone image encoder (\textbf{CN} for CarlaNet \cite{codevilla2018end} or \textbf{RN} for ResNet34), multi-task learning (\textbf{MT}) and loss (\textbf{hLoss} for hard weight loss and \textbf{uLoss} for uncertainty weighted loss). The influence of data augmentation (\textbf{Aug}) is also evaluated. Detailed results of task completeness in test scenarios are provided in Tab.\ref{tab:ablation_task_test}.

Experiments in the first group compares between different backbones and demonstrate that data augmentation is of vital importance to baseline models in our benchmark, especially for ResNet34. Without data augmentation, baseline models have a poor performance due to high timeout rate. Through modeling Lat. and Lon. control as multi-task, performance of models in the second group greatly exceeds that of single-task baselines with respect to success rate and timeout rate.

The last group, which uses uncertainty weighted loss instead of hard weight loss, achieves the best testing performance. Our model adaptively learns to balance between lateral and longitudinal control tasks and further reduces the relatively high collision rates in the second group. 

\section{Conclusion and Future Works}
\label{sec:conclusion}
This work studies DIL-based autonomous control for intersection navigation with pedestrians interaction. We propose a multi-task conditional imitation learning method to adapt both lateral and longitudinal control tasks simultaneously, where task-dependent uncertainties are learned to weight tasks. We applied the presented approach to our proposed IntersectNav benchmark and learned from human demonstrations. Experimental results show that the proposed multi-task learning and uncertainty weighting improves performance a lot. There remains room for progress, where interaction with other vehicles and reaction to traffic signals are left for future work.
%\small
\bibliographystyle{IEEEtran}
	\bibliography{ref.bib}

\begin{thebibliography}{10}
\providecommand{\url}[1]{#1}
\csname url@samestyle\endcsname
\providecommand{\newblock}{\relax}
\providecommand{\bibinfo}[2]{#2}
\providecommand{\BIBentrySTDinterwordspacing}{\spaceskip=0pt\relax}
\providecommand{\BIBentryALTinterwordstretchfactor}{4}
\providecommand{\BIBentryALTinterwordspacing}{\spaceskip=\fontdimen2\font plus
\BIBentryALTinterwordstretchfactor\fontdimen3\font minus
  \fontdimen4\font\relax}
\providecommand{\BIBforeignlanguage}[2]{{%
\expandafter\ifx\csname l@#1\endcsname\relax
\typeout{** WARNING: IEEEtran.bst: No hyphenation pattern has been}%
\typeout{** loaded for the language `#1'. Using the pattern for}%
\typeout{** the default language instead.}%
\else
\language=\csname l@#1\endcsname
\fi
#2}}
\providecommand{\BIBdecl}{\relax}
\BIBdecl

\bibitem{shu2020autonomous}
K.~Shu, H.~Yu, X.~Chen, L.~Chen, Q.~Wang, L.~Li, and D.~Cao, ``Autonomous
  driving at intersections: a critical-turning-point approach for left turns,''
  in \emph{ITSC}, 2020.

\bibitem{shu2021driving}
H.~Shu, T.~Liu, X.~Mu, and D.~Cao, ``Driving tasks transfer using deep
  reinforcement learning for decision-making of autonomous vehicles in
  unsignalized intersection,'' \emph{TVT}, 2021.

\bibitem{li2021planning}
S.~Li, K.~Shu, C.~Chen, and D.~Cao, ``Planning and decision-making for
  connected autonomous vehicles at road intersections: A review,''
  \emph{Chinese Journal of Mechanical Engineering}, 2021.

\bibitem{wei2021autonomous}
L.~Wei, Z.~Li, J.~Gong, C.~Gong, and J.~Li, ``Autonomous driving strategies at
  intersections: Scenarios, state-of-the-art, and future outlooks,'' in
  \emph{ITSC}, 2021.

\bibitem{zhu2021survey}
Z.~Zhu and H.~Zhao, ``A survey of deep rl and il for autonomous driving policy
  learning,'' \emph{TITS}, 2021.

\bibitem{camacho2013model}
E.~F. Camacho and C.~B. Alba, \emph{Model predictive control}.\hskip 1em plus
  0.5em minus 0.4em\relax Springer science \& business media, 2013.

\bibitem{qian2015decentralized}
X.~Qian, J.~Gregoire, A.~De~La~Fortelle, and F.~Moutarde, ``Decentralized model
  predictive control for smooth coordination of automated vehicles at
  intersection,'' in \emph{European control conference (ECC)}, 2015.

\bibitem{schildbach2016collision}
G.~Schildbach, M.~Soppert, and F.~Borrelli, ``A collision avoidance system at
  intersections using robust model predictive control,'' in \emph{IV}, 2016.

\bibitem{misir1996design}
D.~Misir, H.~A. Malki, and G.~Chen, ``Design and analysis of a fuzzy
  proportional-integral-derivative controller,'' \emph{Fuzzy sets and systems},
  1996.

\bibitem{kuutti2020survey}
S.~Kuutti, R.~Bowden, Y.~Jin, P.~Barber, and S.~Fallah, ``A survey of deep
  learning applications to autonomous vehicle control,'' \emph{TITS}, 2020.

\bibitem{bouton2019safe}
M.~Bouton, A.~Nakhaei, K.~Fujimura, and M.~J. Kochenderfer, ``Safe
  reinforcement learning with scene decomposition for navigating complex urban
  environments,'' in \emph{IV}, 2019.

\bibitem{liang2018cirl}
X.~Liang, T.~Wang, L.~Yang, and E.~Xing, ``Cirl: Controllable imitative
  reinforcement learning for vision-based self-driving,'' in \emph{ECCV}, 2018.

\bibitem{zhang2021end}
Z.~Zhang, A.~Liniger, D.~Dai, F.~Yu, and L.~Van~Gool, ``End-to-end urban
  driving by imitating a reinforcement learning coach,'' in \emph{ICCV}, 2021.

\bibitem{bojarski2016end}
M.~Bojarski, D.~Del~Testa, D.~Dworakowski, B.~Firner, B.~Flepp, P.~Goyal, L.~D.
  Jackel, M.~Monfort, U.~Muller, J.~Zhang \emph{et~al.}, ``End to end learning
  for self-driving cars,'' \emph{arXiv:1604.07316}, 2016.

\bibitem{codevilla2018end}
F.~Codevilla, M.~M{\"u}ller, A.~L{\'o}pez, V.~Koltun, and A.~Dosovitskiy,
  ``End-to-end driving via conditional imitation learning,'' in \emph{ICRA},
  2018.

\bibitem{codevilla2019exploring}
F.~Codevilla, E.~Santana, A.~M. L{\'o}pez, and A.~Gaidon, ``Exploring the
  limitations of behavior cloning for autonomous driving,'' in \emph{ICCV},
  2019.

\bibitem{zhao2019sam}
A.~Zhao, T.~He, Y.~Liang, H.~Huang, G.~V.~d. Broeck, and S.~Soatto, ``Sam:
  Squeeze-and-mimic networks for conditional visual driving policy learning,''
  \emph{arXiv:1912.02973}, 2019.

\bibitem{chen2020learning}
D.~Chen, B.~Zhou, V.~Koltun, and P.~Kr{\"a}henb{\"u}hl, ``Learning by
  cheating,'' in \emph{CoRL}, 2020.

\bibitem{ross2010efficient}
S.~Ross and D.~Bagnell, ``Efficient reductions for imitation learning,'' in
  \emph{Proceedings of the International Conference on Artificial Intelligence
  and Statistics}, 2010.

\bibitem{de2019causal}
P.~de~Haan, D.~Jayaraman, and S.~Levine, ``Causal confusion in imitation
  learning,'' \emph{NIPS}, 2019.

\bibitem{sauer2018conditional}
A.~Sauer, N.~Savinov, and A.~Geiger, ``Conditional affordance learning for
  driving in urban environments,'' in \emph{CoRL}, 2018.

\bibitem{dosovitskiy2017carla}
A.~Dosovitskiy, G.~Ros, F.~Codevilla, A.~Lopez, and V.~Koltun, ``Carla: An open
  urban driving simulator,'' in \emph{CoRL}, 2017.

\bibitem{carlaleaderboard}
C.~team, ``Carla autonomous driving leaderboard,''
  \url{https://leaderboard.carla.org}, CARLA team.

\bibitem{chen2015deepdriving}
C.~Chen, A.~Seff, A.~Kornhauser, and J.~Xiao, ``Deepdriving: Learning
  affordance for direct perception in autonomous driving,'' in \emph{ICCV},
  2015.

\bibitem{pomerleau1988alvinn}
D.~Pomerleau, ``{ALVINN:} an autonomous land vehicle in a neural network,'' in
  \emph{NIPS}, 1988.

\bibitem{prakash2020exploring}
A.~Prakash, A.~Behl, E.~Ohn-Bar, K.~Chitta, and A.~Geiger, ``Exploring data
  aggregation in policy learning for vision-based urban autonomous driving,''
  in \emph{CVPR}, 2020.

\bibitem{ohn2020learning}
E.~Ohn-Bar, A.~Prakash, A.~Behl, K.~Chitta, and A.~Geiger, ``Learning
  situational driving,'' in \emph{CVPR}, 2020.

\bibitem{ruder2017overview}
S.~Ruder, ``An overview of multi-task learning in deep neural networks,''
  \emph{arXiv:1706.05098}, 2017.

\bibitem{zhang2021survey}
Y.~Zhang and Q.~Yang, ``A survey on multi-task learning,'' \emph{IEEE
  Transactions on Knowledge and Data Engineering}, 2021.

\bibitem{liao2016understand}
Y.~Liao, S.~Kodagoda, Y.~Wang, L.~Shi, and Y.~Liu, ``Understand scene
  categories by objects: A semantic regularized scene classifier using
  convolutional neural networks,'' in \emph{ICRA}, 2016.

\bibitem{sermanet2013overfeat}
P.~Sermanet, D.~Eigen, X.~Zhang, M.~Mathieu, R.~Fergus, and Y.~LeCun,
  ``Overfeat: Integrated recognition, localization and detection using
  convolutional networks,'' \emph{arXiv:1312.6229}, 2013.

\bibitem{eigen2015predicting}
D.~Eigen and R.~Fergus, ``Predicting depth, surface normals and semantic labels
  with a common multi-scale convolutional architecture,'' in \emph{ICCV}, 2015.

\bibitem{xu2018shared}
J.~Xu, Q.~Liu, H.~Guo, A.~Kageza, S.~AlQarni, and S.~Wu, ``Shared multi-task
  imitation learning for indoor self-navigation,'' in \emph{IEEE Global
  Communications Conference}, 2018.

\bibitem{xu2017end}
H.~Xu, Y.~Gao, F.~Yu, and T.~Darrell, ``End-to-end learning of driving models
  from large-scale video datasets,'' in \emph{CVPR}, 2017.

\bibitem{kim2020multi}
I.~Kim, H.~Lee, J.~Lee, E.~Lee, and D.~Kim, ``Multi-task learning with future
  states for vision-based autonomous driving,'' in \emph{ACCV}, 2020.

\bibitem{ishihara2021multi}
K.~Ishihara, A.~Kanervisto, J.~Miura, and V.~Hautamaki, ``Multi-task learning
  with attention for end-to-end autonomous driving,'' in \emph{CVPR}, 2021.

\bibitem{kendall2018multi}
A.~Kendall, Y.~Gal, and R.~Cipolla, ``Multi-task learning using uncertainty to
  weigh losses for scene geometry and semantics,'' in \emph{CVPR}, 2018.

\bibitem{he2016deep}
K.~He, X.~Zhang, S.~Ren, and J.~Sun, ``Deep residual learning for image
  recognition,'' in \emph{CVPR}, 2016.

\bibitem{kingma2014adam}
D.~P. Kingma and J.~Ba, ``Adam: A method for stochastic optimization,''
  \emph{arXiv:1412.6980}, 2014.

\end{thebibliography}
\end{document}